\documentclass[conference]{IEEEtran}
\usepackage{blindtext, graphicx}

\ifCLASSINFOpdf
\else
\fi

\usepackage[utf8]{inputenc}
\usepackage{multirow}

\usepackage{ifpdf}
\usepackage{epstopdf}
\DeclareGraphicsExtensions{.eps}
\usepackage[font=footnotesize]{subfig}
\usepackage{caption}
\usepackage{comment}
\usepackage{amsmath,amsthm,amsfonts,amssymb}
\usepackage{cite}
\usepackage{algorithm, algorithmic}
\usepackage{graphics}
\usepackage{color}

\IEEEoverridecommandlockouts

\begin{document}
\title{Submodular Load Clustering with Robust \\ Principal Component Analysis} 

\author{	
\IEEEauthorblockN{
                Yishen~Wang$^{1}$,
                Xiao~Lu$^{2}$,
                Yiran~Xu$^{3}$,
                Di~Shi$^{1}$,
                Zhehan~Yi$^{1}$,
                Jiajun~Duan$^{1}$,
				Zhiwei~Wang$^{1}$
}
\IEEEauthorblockA{$^{1}$GEIRI North America, San Jose, CA, USA}
\IEEEauthorblockA{$^{2}$State Grid Jiangsu Electric Power Company Ltd., Nanjing, Jiangsu, China}
\IEEEauthorblockA{$^{3}$State Grid Nanjing Power Supply Company, Nanjing, Jiangsu, China}
\IEEEauthorblockA{Email: yishen.wang@geirina.net}
\thanks{This work is funded by SGCC Science and Technology Program under contract no. SGSDYT00FCJS1700676.
}
\vspace{-25pt}
}

\maketitle

\begin{abstract}
Traditional load analysis is facing challenges with the new electricity usage patterns due to demand response as well as increasing deployment of distributed generations, including photovoltaics (PV), electric vehicles (EV), and energy storage systems (ESS). At the transmission system, despite of irregular load behaviors at different areas, highly aggregated load shapes still share similar characteristics. Load clustering is to discover such intrinsic patterns and provide useful information to other load applications, such as load forecasting and load modeling. This paper proposes an efficient submodular load clustering method for transmission-level load areas. Robust principal component analysis (R-PCA) firstly decomposes the annual load profiles into low-rank components and sparse components to extract key features. A novel submodular cluster center selection technique is then applied to determine the optimal cluster centers through constructed similarity graph. Following the selection results, load areas are efficiently assigned to different clusters for further load analysis and applications. Numerical results obtained from PJM load demonstrate the effectiveness of the proposed approach.
\end{abstract}

\begin{IEEEkeywords}
Clustering, Load, Machine learning, Robust principal component analysis, Submodular optimization
\end{IEEEkeywords}

\IEEEpeerreviewmaketitle

\section{Introduction}

Load has been one of the fundamental parts of power systems since the systems exist. Conventionally, load components include electric heaters, Heating, Ventilation, and Air Conditioning (HVAC), household electric appliances (refrigerators, TV, washer, dryer, dishwasher, etc.), lighting load, induction motors (industrial and agricultural), industrial facilities \cite{kundur_1994_book}. 

From the perspective of transmission system operators, load bus represents the aggregated load for all the downstream loads within the sub-transmission and distribution systems \cite{kundur_1994_book}. Currently, there is an increasing trend to deploy distributed resources in the system, such as PV \cite{zhehan_2017_fault}, demand response \cite{desong_2015_human}, EV \cite{zhe_2016_dr}, and energy storage \cite{yishen_2017_lookahead}. In addition, AC microgrids \cite{xiaohu_2017_mg}, DC microgrids \cite{jiajun_2018_decentral} or hybrid AC/DC microgrids \cite{yishen_2018_hybrid, zhehan_2018_unified} can also be viewed as aggregated loads. Under this paradigm, originally passive load is becoming more and more active, which places new challenges on load analysis for system operation and planning.

In power systems, there are various load analysis applications, including load forecasting \cite{hong_2014_long}, load modeling \cite{chang_2018_bayesian, yachen_2017_enhancment}, load disaggregation \cite{yongli_2014_load}, and load clustering \cite{chicco_2006_comparison, liran_2017_loadcharacter, mets_2016_twostage, yi_2015_review, yang_2017_kshape}. Within these applications, load clustering serves as an effective intermediate step to improve the performances of others. Considering the intrinsic load pattern similarities, it is not practical and not necessary to fit every single load bus with a unique model in a realistically-sized system. Loads within the same cluster can share the same parameter set for modeling and forecasting. In addition, as load behavior randomness always exists, load clustering helps load modeling and load forecasting to learn a more generalized model.

Chicco \emph{et. al} \cite{chicco_2006_comparison} aim to group similar customer consumption behaviors and compare several unsupervised methods, including hierarchical clustering and $K$-Means for clustering as well as Sammon map and principal component analysis (PCA) for dimensionality reduction. Li \emph{et. al} \cite{liran_2017_loadcharacter} propose to decompose smart meter data in the spectral domain and apply discrete Fourier transform (DFT) and discrete wavelet transform (DWT) for load characterization at different aggregation levels. Mets \emph{et. al} \cite{mets_2016_twostage} similarly adopt a two-stage method to analyze and identify load patterns through fast wavelet transformation (FWT). Wang \emph{et. al} \cite{yi_2015_review} conduct a detailed review to discuss various load profiling methods and their applications. Yang \emph{et. al} \cite{yang_2017_kshape} apply $K$-Shape method to cluster building load patterns and demonstrate the improved forecasting accuracy with such clustering results.   

A large number of papers discuss how load profiles can be clustered with various clustering methods, \emph{e.g.} $K$-Means \cite{chicco_2006_comparison} and $K$-Shape \cite{yang_2017_kshape}. However, previous work mainly focused on low-voltage-level smart meter data, while this paper evaluates the clustering method at load areas for transmission system applications. This paper makes the following contributions: 
\begin{itemize}
    \item A robust principal component analysis (R-PCA) is applied to decompose the load time series into low-rank and sparse components to extract key features. In addition, R-PCA also effectively mitigates data quality issues caused by corrupted data or missing values. 
    \item A novel submodular selection technique is proposed to determine cluster centers. Load areas are ranked in order as the center candidates, where higher order indicates higher priority to be chosen as the cluster centers. According to such rank, without repeating the whole clustering process, a different total cluster number $K$ simply requires picking $K$ first candidates and re-assigning load profiles. This greatly improves the overall clustering efficiency. Unlike $K$-Means, this method is deterministic, so the clustering results are stable without randomness. 
    \item A detailed case study based on PJM load data demonstrates the effectiveness of the proposed method in clustering highly aggregated transmission load areas. 
\end{itemize}

The rest of the paper is organized as follows. Section II presents the detailed method to apply robust PCA and submodular clustering method for partitioning load areas into different groups. Section III provides the numerical results based on the PJM load area data. Section IV concludes the paper.

\section{Submodular Load Clustering Method}
This section illustrates the proposed submodular clustering method in details.

\subsection{Data Normalization}
After collecting $N^I$ load area data from the database, each load area $i$ is represented with a column vector $x_i \in \mathcal{R}^{N^{T}}$ recording historic annual load profiles with $N^T$ measurements. Different load areas typically own different peak load $X_{i}^{max}$, which ranges from 200 MW to 20 GW for PJM \cite{pjm_load}. For computation stability and ease of comparison, these profiles are normalized with feature scaling to have scales between 0 and 1 as shown in equation~\eqref{eqref:normalization}. Then, a load data matrix $M$ with size $N^{T} \times N^{I}$ is generated. Each column  $y_i \in \mathcal{R}^{N^{T}}$ represents a normalized load profile.  

\begin{equation} \label{eqref:normalization}
    y_i = \frac{x_i - X_{i}^{min}}{X_{i}^{max} - X_{i}^{min}} 
\end{equation}

\subsection{Robust Principal Component Analysis}
Principal component analysis (PCA) has been widely used to reduce the data dimension and extract features \cite{yi_2015_review}. Conventionally, this PCA can be formulated with the optimization form as shown in equations~\eqref{eqref:pca}. The goal is to find a rank-$k$ component $L$ to minimize the $\ell_2$-norm of reconstruction errors between the original data $M$ and low-rank component $L$. Singular value decomposition (SVD) is commonly used for solving this problem.

\begin{equation} \label{eqref:pca}
    \begin{aligned}
        & \text{minimize}    & \quad \quad   &  \left\lVert M - L  \right\rVert \\
        & \text{subject to}  & \quad \quad   & \text{rank}(L) \leq k
    \end{aligned}
\end{equation}

However, PCA cannot perform well when the data is not thoroughly cleaned. Corrupted measurements could lead to poor reduction results even with single grossly corrupted errors. Since real-world applications always come with data quality issues, a robust version is necessary to extend the existing PCA work. As suggested in \cite{Candes_2011_rpca}, Robust PCA (R-PCA) is proposed to robustify conventional PCA. In power systems, R-PCA has also been applied in the data cleaning \cite{mateos_2013_load} and false data injection. 

In this paper, R-PCA serves as the filter to decompose the normalized load data into low-rank and sparse components, whose information will be further extracted. Even though PCA itself is a linear mapping to transform data, keeping both components in this robust version still withholds as much information as possible, and suits for load time series analysis.    

The Principal Component Pursuit (PCP) form of R-PCA is shown in equations~\eqref{eqref:rpca}, whose objective is to minimize the weighted sum of the nuclear-norm of low-rank matrix $L$ and $\ell_1$-norm of sparse matrix $S$, subject to the original matrix condition. Nuclear norm computes the sum of singular values of matrix as in \eqref{eqref:nuclearnorm}, and $\ell_1$-norm enforces sparsity for the matrix $S$ as in \eqref{eqref:l1norm}. 
\begin{equation} \label{eqref:rpca}
    \begin{aligned}
        & \text{minimize}    & \quad \quad   & \left\lVert L \right\rVert_{*} + \mu \left\lVert S \right\rVert_{1} \\
        & \text{subject to}  & \quad \quad   & L + S = M
    \end{aligned}
\end{equation}
\begin{equation} \label{eqref:nuclearnorm}
    \left\lVert L \right\rVert_{*} = \sum_{i} \sigma_{i} (L)
\end{equation}
\begin{equation} \label{eqref:l1norm}
    \left\lVert S \right\rVert_{1} = \sum_{ij} |S_{ij}|
\end{equation}
Another unique and remarkable feature for this formulation is that no tuning parameter is required. Weighting factor $\mu$ is theoretically determined with equation \eqref{eqref:mu} under mild conditions.
\begin{equation} \label{eqref:mu}
    \mu = \frac{1}{\sqrt{\max(N^T, N^I)}}
\end{equation}
This tractable convex optimization can be solved to recover the original data matrix efficiently and exactly in the PCP form. In addition, algorithms like iterative thresholding, accelerated proximal gradient, augmented Lagrangian multipliers can also be applied to solve this problem. Further details and proofs should be referred to \cite{Candes_2011_rpca}.

\subsection{Feature Extraction}
With decomposed low-rank and sparse components from the normalized load profiles, load area feature vector length is actually doubled and quite long. The total feature number reaches $2 \times N^{T}$ for each area. For hourly metered load, annual feature numbers are $2 \times 8760 = 17520$. For smart meter data or Phasor Measurement Unit (PMU) data with much higher sampling rate, the total feature length is even longer. Directly feeding these long vectors for clustering does not help machine learning perform better. Instead, due to ``curse of dimensionality'', the algorithm may even perform worse. Therefore, intrinsic and representative load features needs to be extracted. 

As mentioned in \cite{yi_2015_review}, one common feature extraction approach is through dimensionality reduction techniques like PCA or autoencoders. However, the number of features to be selected and the method to be used is still a data-dependent question.

\begin{figure}[htb]
	\includegraphics[scale=0.6]{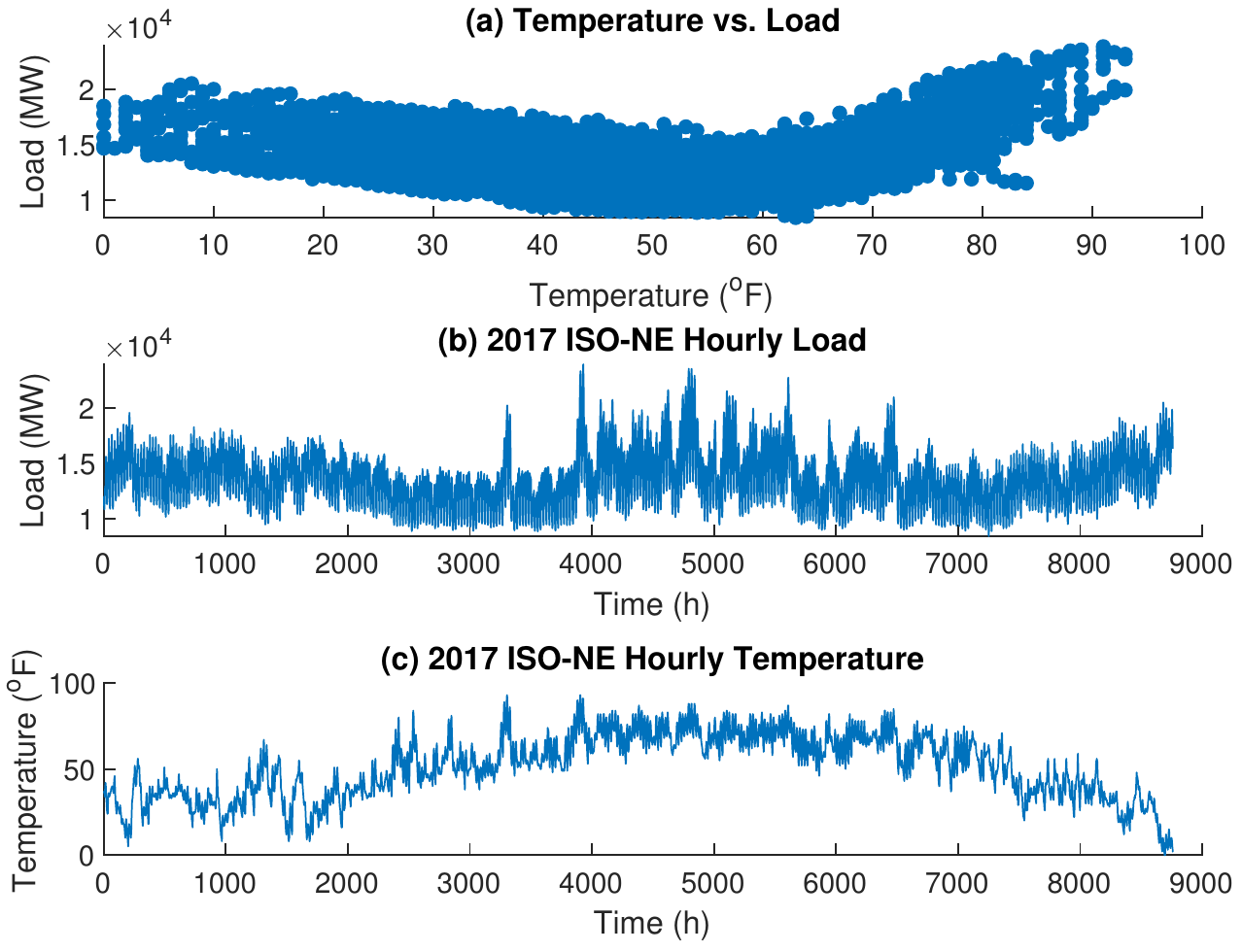}
	\centering
	\caption{2017 ISO New England Load and Temperature} 
	\label{plot:isone}
\end{figure}

Load time series are highly weather dependent \cite{hong_2014_long}, especially with temperatures. As shown in Fig.~\ref{plot:isone}(a) ISO New England temperature and load plot, temperature has a strong quadratic relationship with the load. In addition, Fig.~\ref{plot:isone}(b) and (c) also indicate a clear seasonal trend for both temperature and load. This seasonal load pattern difference motivates to capture such information through designed features. Rather than automatically extract features without clear physical meanings, a feature engineering approach is developed based on the understanding of the load data. 

Following the convention in ISO New England load data \cite{isone_load}, the whole year is split into winter and summer two seasons. The summer period is June to September, while the winter period is October to May. Even though this is the definition for ISO New England, the same principles can be applied to PJM as well. 

For each season, we extract the seasonal average load, seasonal load standard deviation, seasonal maximum load and seasonal minimum load to form the area load feature set $z_i$. Due to the decomposed low-rank and sparse components from R-PCA, the total feature length for each area is $2\text{ (matrix component)} \times 2\text{ (season)} \times 4\text{ (feature)} = 16$. 

\vspace{-10pt}

\subsection{Similarity Graph Construction}

Similarity graph is constructed through the similarity matrix $W$ and corresponding $w_{ij}$ for each load area pair. Distance metric with $\ell_2$-norm and similarity metric with radial basis kernel function (RBF) are computed with equations \eqref{eqref:l2} and \eqref{eqref:rbf}. 
\begin{equation} \label{eqref:l2}
d_{ij} = \left\lVert z_i - z_j \right\rVert_{2}
\end{equation}
\begin{equation} \label{eqref:rbf}
w_{ij} = e^{- \frac{d_{ij}}{\lambda}}
\end{equation}
where $z_i$ and $z_j$ are the load features for area $i$ and $j$. Parameter $\lambda$ controls the similarity scaling. Typically, tuning $\lambda$ is based on pairwise $\ell_2$-norm distance distribution values.

\subsection{Submodular Cluster Center Selection}
Submodular optimization has shown great success in providing computationally efficient and theoretically-bounded solutions to feature selection, training data selection and other machine learning applications \cite{yuzong_2013_submodular}. In power systems, such submodular optimization concept also presents satisfactory results in scenario reduction \cite{yishen_2017_scenred}, PMU placement and storage siting. 

\textit{\textbf{Diminishing return property}}: For every $X, Y \subseteq \Omega$ with $X \subseteq Y$ and every $x \in \Omega \setminus Y$, we have that $f(X \cup \{x\}) - f(X) \geq f(Y \cup \{x\}) - f(Y)$.

Diminishing return property is one of the most important properties for submodular optimization. It states that the incremental gain from selecting one more element is always non-increasing. Therefore, this incremental gain has an upper bound which helps to derive efficient algorithms \cite{minoux_1978_accelerated}. Nemhauser \cite{nemhauser_1978_analysis} prove that a simple greedy algorithm can solve the submodular function optimization with an optimality-bounded solution when it is monotone non-decreasing.

Motivated by the classic facility location problem, load clustering can be formulated as a submodular cluster center selection problem. Given $N^I$ load areas, the task is to find $K$ load cluster centers to represent all $N^I$ load areas. From set function perspective, given original set $\Omega$, load cluster center selection tries to find a selected set $\Gamma$ to maximize the set-wise similarities between the original set $\Omega$ and selected set $\Gamma$, subject to the cardinality constraint. This has been formulated in discrete optimization problem~\eqref{eqref:submodular}.
\begin{equation} \label{eqref:submodular}
    \begin{aligned}
        & f(\Gamma)  & = & \quad \quad  \max_{i \in \Omega} \quad (\sum_{i \in \Omega} \max_{j \in \Gamma} w_{ij}) \\
        & \text{subject to} &  & \quad \quad  card(\Gamma) \leq K
    \end{aligned}
\end{equation}

This formulation follows the facility location problem convention, so it is submodular with proof. As the objective is monotone non-decreasing, nice computational properties are applied to derive accelerated fast greedy algorithm \cite{minoux_1978_accelerated, yishen_2017_scenred}. Algorithm~\ref{algo:accgreedy} presents the detailed algorithm to solve the load cluster center selection problem with guaranteed solution quality.

\begin{algorithm}[hpbt] 
	\caption{Submodular cluster center selection algorithm} \label{algo:accgreedy}
	\begin{algorithmic}[1]
		\STATE{\bf{Initialize} $\Gamma \leftarrow \emptyset$, ${f(\Gamma)} \leftarrow 0$}, 
		
		\FOR{$ i = 1,2,\cdots, N^{I}$}
		\STATE{$v_i \leftarrow {f(i)}$}
		\ENDFOR
		
		\WHILE{${card(\Gamma)}\le K$}
		\STATE{$j \leftarrow arg \max_{i \in \Omega \setminus \Gamma} v_i$}
		\STATE{$\delta \leftarrow f(\Gamma \cup \{ j \}) - f(\Gamma)$}
		\STATE{$v_j \leftarrow \delta$}
		
		\IF{$\delta > \max_{i \in \Omega \setminus (\Gamma \cup j)} v_i$}
		\STATE{$\Gamma \leftarrow \Gamma \cup \{ j \}$}	\COMMENT{Diminishing return property}
		\STATE{$v_j \leftarrow 0$} 
		\ELSIF{$ \delta \le 0$}
		\STATE {\bf{Break}} \COMMENT {Solution is (locally) optimal}
		\ENDIF
		\ENDWHILE
		\STATE{return $\Gamma$}		
	\end{algorithmic}
\end{algorithm}

Basically, this accelerated algorithm firstly initializes each set element. Then, these initialized values serve as the upper bounds to efficiently select new set elements. Even though it is greedy, the solution is always optimal or near-optimal. As the objective is monotone non-decreasing, the algorithm will stop when the set cardinality $K$ is met. In other words, the algorithm will finish when it finds $K$ best cluster centers. 

The $K$ cluster centers are ranked with the selection order. Higher order suggests a higher priority to be chosen as the cluster center to represent remaining data points.

\subsection{Load Cluster Assignment}
After obtaining the cluster center rank list or rank set $\Gamma$, arbitrary $K$-cluster can be determined through selecting first $K$ load areas in the list as the cluster centers. Then, the rest load areas are assigned to the load area cluster centers with highest pairwise similarity as in equation \eqref{eqref:cluster}.
\begin{equation} \label{eqref:cluster}
    c_i = \max_{j \in \Gamma} w_{ij}
\end{equation}
By setting this $K$ with a large value, any cluster number less than $K$ is efficiently computed and assigned without repeating the entire clustering process. For example, to compare clustering results with the clustering number from 1 to $K$, the proposed algorithm only conducts 1-time cluster center selection and $K$-time cluster assignments, whereas $K$-Means needs to perform $K$-time clusterings and $K$-time cluster assignments. The computation efforts are greatly reduced especially when there are large number of data points to be clustered. 

In addition, $K$-Means results heavily rely on the random initialization, so it generally needs to repeat several times to find the best clustering partition. On the contrary, the proposed entire process is deterministic without randomness, and it has good interpretabilities due to this rank list as well.

\begin{figure}[htb]
	\includegraphics[scale=0.35]{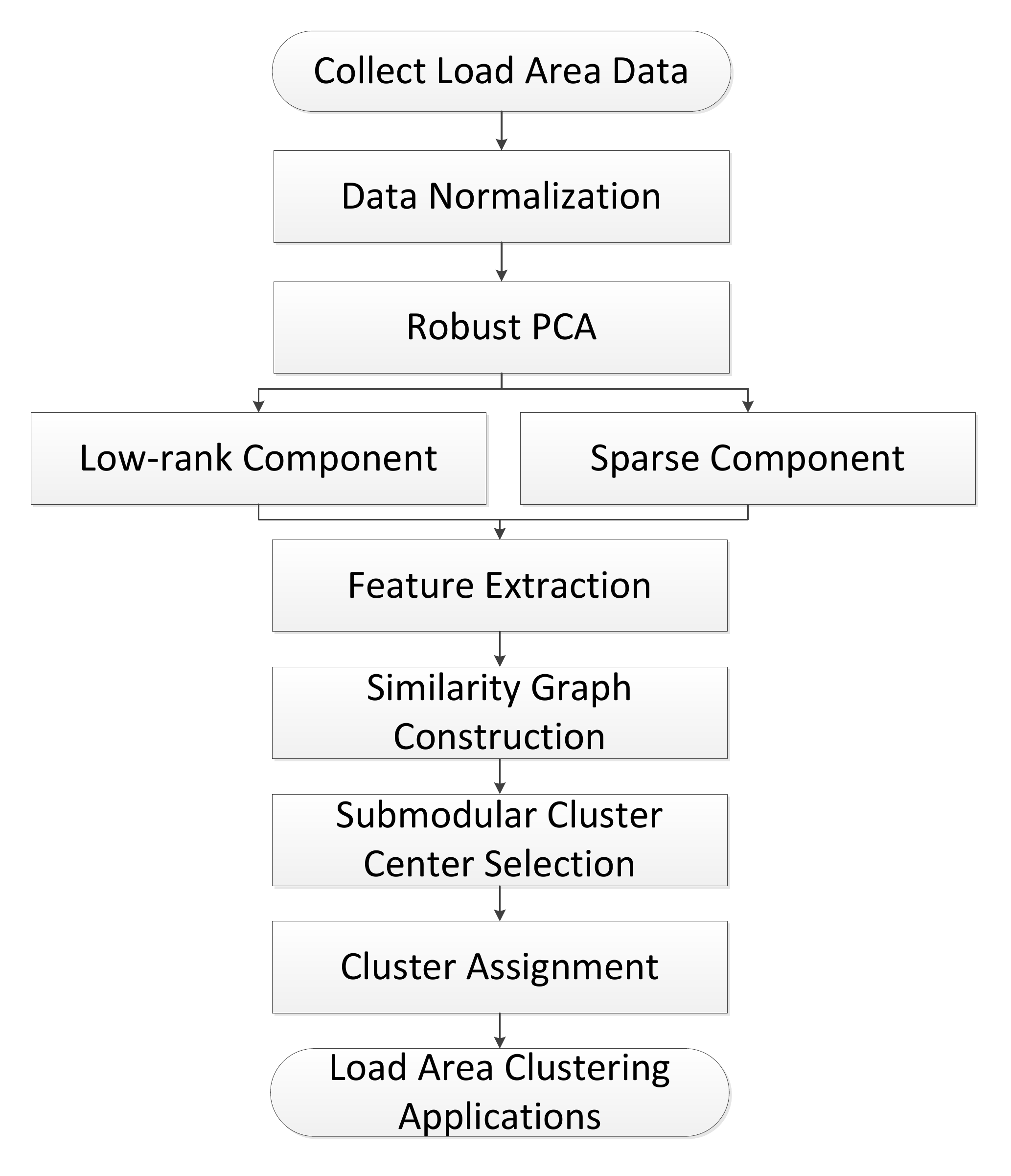}
	\centering
	\caption{Submodular-based Clustering Flowchart} 
	\label{plot:flowchart}
\end{figure}

The overall framework of the submodular load clustering method is presented in Fig.~\ref{plot:flowchart}.

\vspace{-10pt}

\section{Case Study}
\subsection{Simulation Setup}
The proposed submodular load clustering method are tested using PJM area load data \cite{pjm_load}. In this dataset, 27 load areas represent 27 fully metered electric distribution companies in the PJM territories. Year 2017 annual load profiles are chosen to represent most up-to-date load shapes due to the increasing deployment of distributed energy resources. All simulations are carried out in MATLAB 2016a on a desktop with Intel Core 4.00 GHz processor and 12.0 GB of RAM.

\subsection{Numerical Results}

\begin{figure}[htb]
	\includegraphics[scale=0.45]{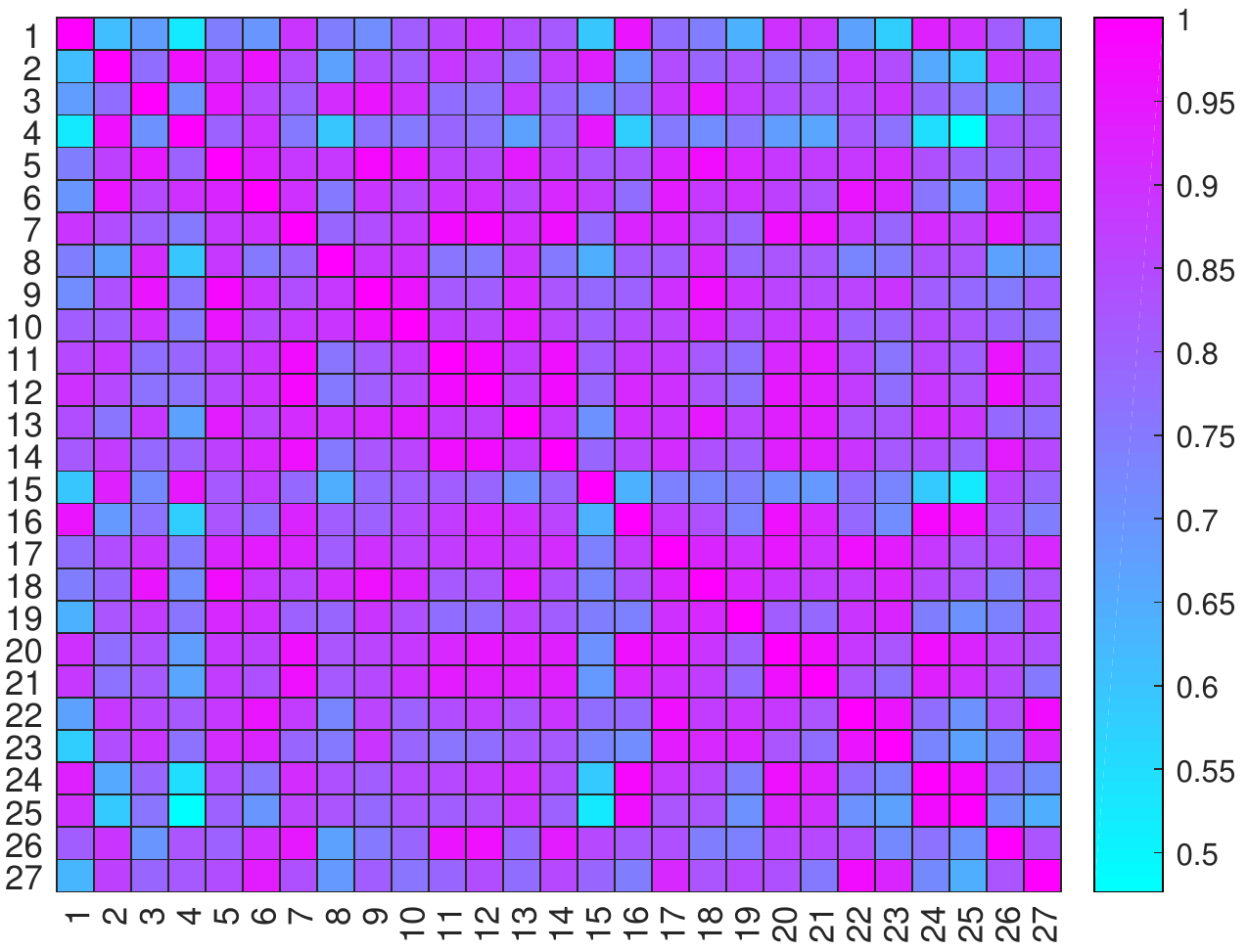}
	\centering
	\caption{PJM Load Area Heatmap} 
	\label{plot:heatmap}
\end{figure}

Fig.~\ref{plot:heatmap} shows the heatmap for 27 PJM load areas. Since each load area covers a distribution company, the load aggregation level is very high. It is not surprising to observe high similarities among most areas. However, several areas still show low correlations between each other indicating the existence of multiple clusters.
\vspace{-5pt}
\vspace{-5pt}

\begin{figure}[htb]
	\includegraphics[scale=0.55]{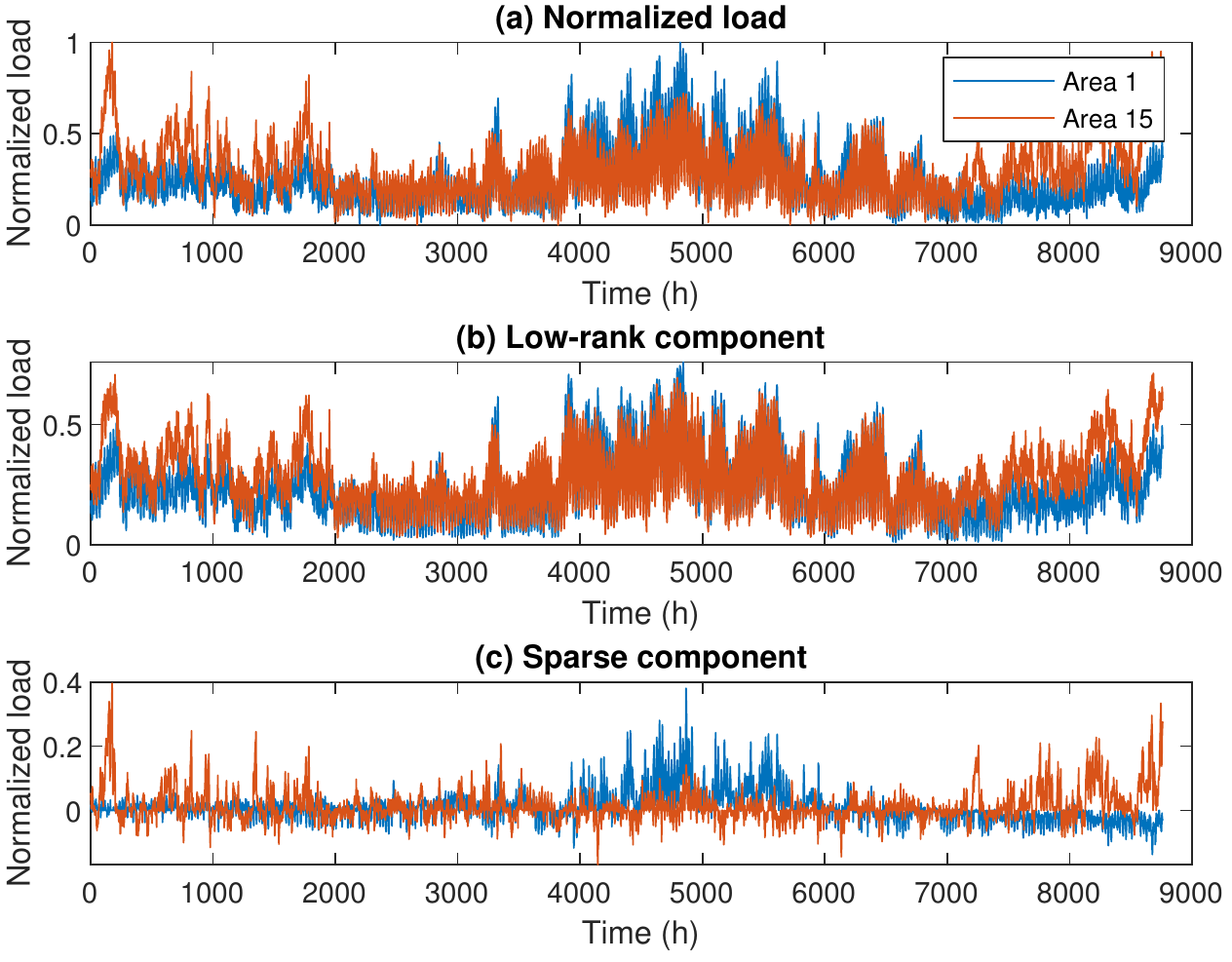}
	\centering
	\caption{Robust PCA Examples} 
	\label{plot:rpca}
\end{figure}

\begin{figure}[htb]
	\includegraphics[scale=0.6]{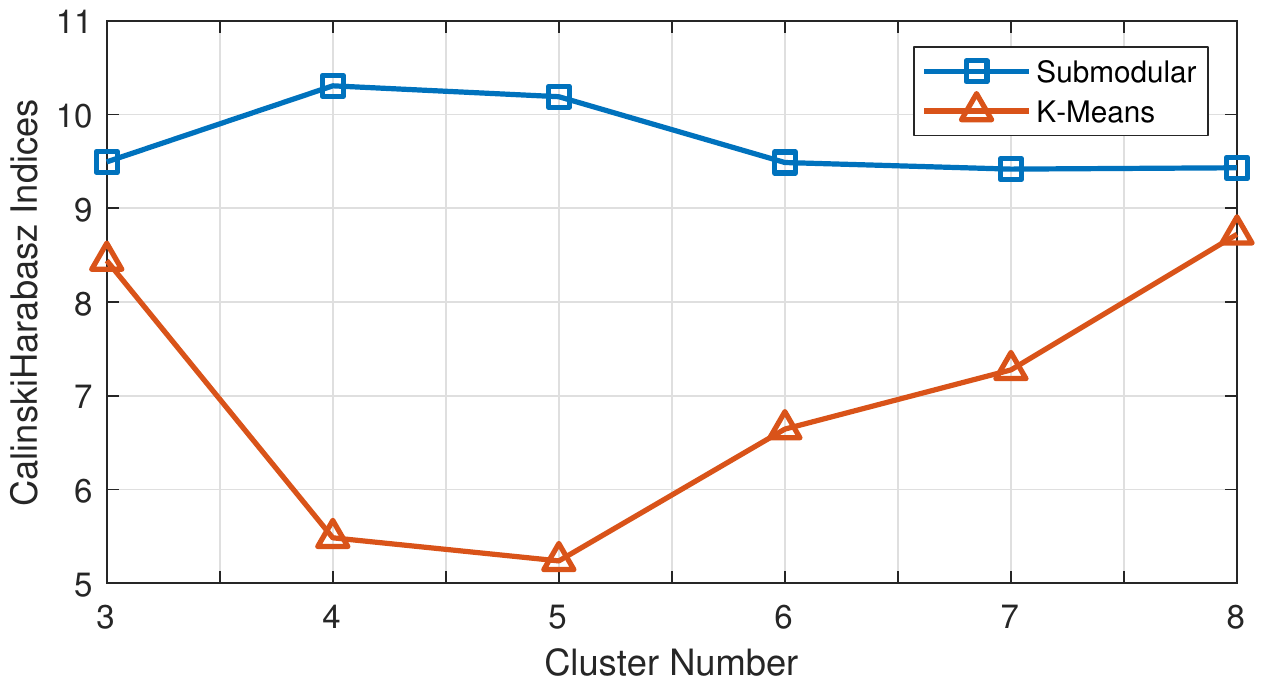}
	\centering
	\caption{Clustering Evaluation} 
	\label{plot:eval}
	\vspace{-15pt}
\end{figure}

Fig.~\ref{plot:rpca} shows one example for the robust PCA and feature extraction. Area 1 and Area 15 both have high summer peaks, whereas area 15 has higher winter peaks. Even though normalized profiles still show such difference, through R-PCA, the sparse components clearly distinguish such pattern difference and provide useful features for clustering.  
\vspace{-5pt}
Fig.~\ref{plot:eval} presents the clustering evaluation with Calinski-Harabasz index. Higher values indicate better clustering results. Compared to $K$-Means, the proposed submodular method reaches better results, and the optimal $K$ is 4.

\begin{figure}[htb]
	\includegraphics[scale=0.65]{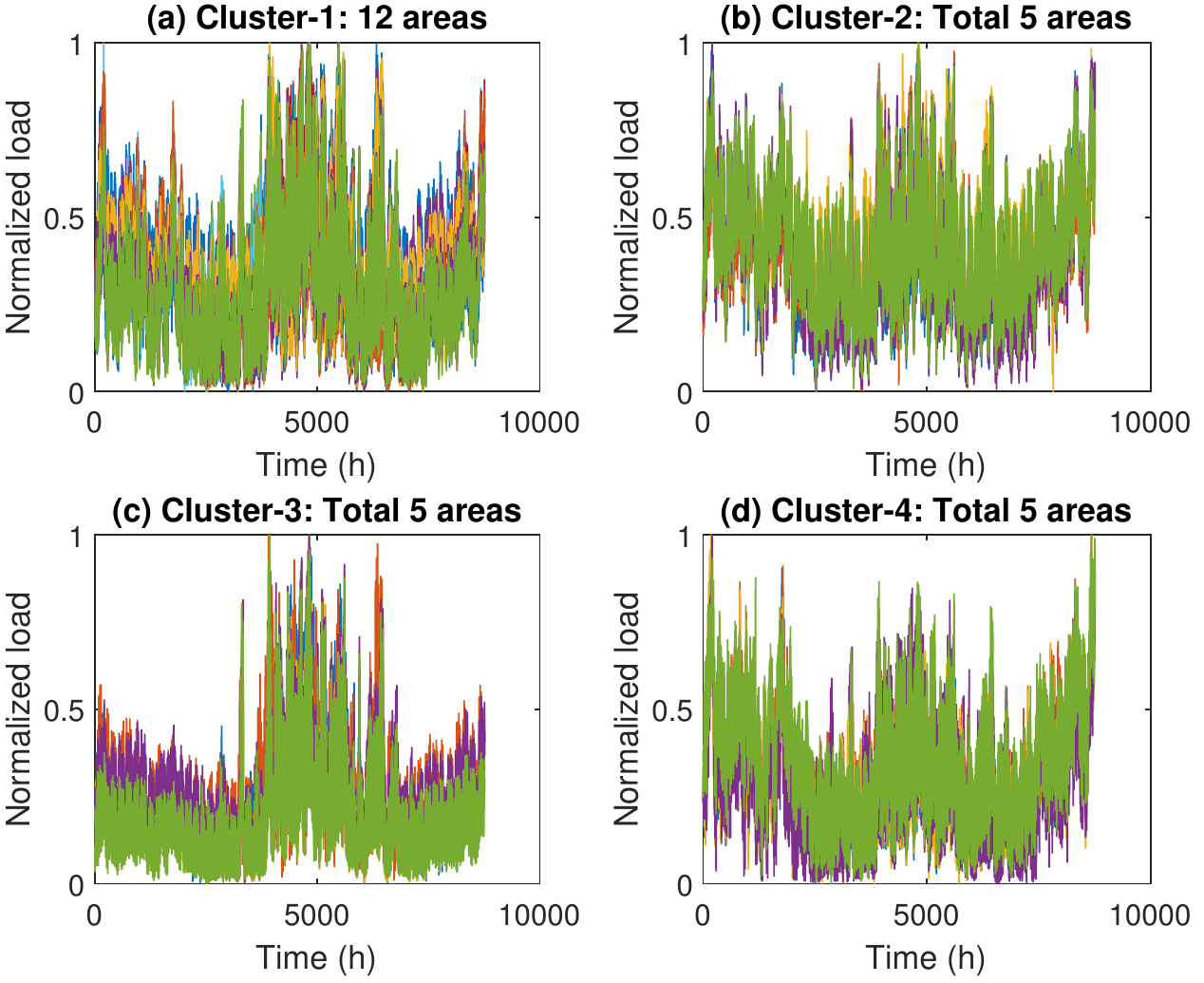}
	\centering
	\caption{Clustering Results for $K$ = 4} 
	\label{plot:cluster}
\end{figure}

Fig.~\ref{plot:cluster} shows the clustering results for 4 load groups. Cluster-1 has high summer peak and winter peak with medium level winter average. Cluster-2 is with relatively similar scale in winter and summer peaks, but higher winter average values than Cluster-1. Cluster-3 shows high summer peak and low winter peak. Cluster-4 has medium-high summer peak and high winter peak. Compared with other clusters, Cluster-1 includes the most load areas. This figure demonstrates the clustering capabilities with the proposed submodular clustering method to extract representative load features.

\vspace{-5pt}

\section{Conclusion}
In this paper, a submodular load clustering method is proposed to efficiently partition the transmission load areas into several groups for better load analysis applications, especially load modeling or forecasting. Robust principal component analysis is applied to extract low-rank and sparse components from normalized load profiles as well as to mitigate corrupted data quality issues. After extracting representative seasonal features, a submodular cluster center selection technique is proposed to efficiently rank the load areas as cluster center candidates. Through scanning the ranked list, different number of clusters can be assigned and evaluated without incurring the clustering process repeatedly. In addition, the proposed method provides deterministic clustering results without randomness. Numerical results from PJM load demonstrate the effectiveness of the proposed method on clustering real transmission-level load areas.


\end{document}